\pdfoutput=1
\documentclass[conference]{IEEEtran}
\usepackage{booktabs}
\usepackage{graphicx}
\usepackage{amsmath}
\usepackage{algorithm}
\usepackage{algorithmicx}  
\usepackage{algpseudocode}  
\usepackage{threeparttable}
\usepackage{textcomp}
\usepackage{multirow}
\usepackage{cite}
\usepackage[numbers,sort&compress]{natbib}
\usepackage{color}

\begin{document}
\title{FedSup: A Communication-Efficient Federated Learning Fatigue Driving Behaviors Supervision Framework}

\author{Chen Zhao, Zhipeng Gao, Qian Wang, Kaile Xiao, Zijia Mo}
\author{Chen~Zhao,~\IEEEmembership{Student~Member,~IEEE,}
Zhipeng~Gao,~\IEEEmembership{Member,~IEEE,}
Qian~Wang,~\IEEEmembership{Student~Member,~IEEE}
Kaile~Xiao,~\IEEEmembership{Student~Member,~IEEE,}
Zijia~Mo,~\IEEEmembership{Student~Member,~IEEE,}
M.~Jamal~Deen,~\IEEEmembership{Fellow,~IEEE,}

}

\maketitle

\begin{abstract}
  With the proliferation of edge smart devices and the Internet of Vehicles (IoV) technologies, intelligent fatigue detection has become one of the most-used methods in our daily driving. To improve the performance of the detection model, a series of techniques have been developed. However, existing work still leaves much to be desired, such as privacy disclosure and communication cost. To address these issues, we propose FedSup, a client-edge-cloud framework for privacy and efficient fatigue detection. Inspired by the federated learning technique, FedSup intelligently utilizes the collaboration between client, edge, and cloud server to realizing dynamic model optimization while protecting edge data privacy. Moreover, to reduce the unnecessary system communication overhead, we further propose a Bayesian convolutional neural network (BCNN) approximation strategy on the clients and an uncertainty weighted aggregation algorithm on the cloud to enhance the central model training efficiency. Extensive experiments demonstrate that the FedSup framework is suitable for IoV scenarios and outperforms other mainstream methods.

\end{abstract}


\IEEEpeerreviewmaketitle

\section{Introduction}
\IEEEPARstart{T}{he} traffic accident is an important issue in driving safety, and according to the official report \cite{refer1}, causes of traffic accidents are generally attributed to drivers, road conditions, and weather, among which abnormal driving behaviors of drivers account for 95\% responsibilities. Hence, how to avoid abnormal driving behaviors during the driving process to ensure the safety of drivers and passengers is an urgent problem.

\begin{figure}[htbp]
  \centering
  \scalebox{0.35}{\includegraphics{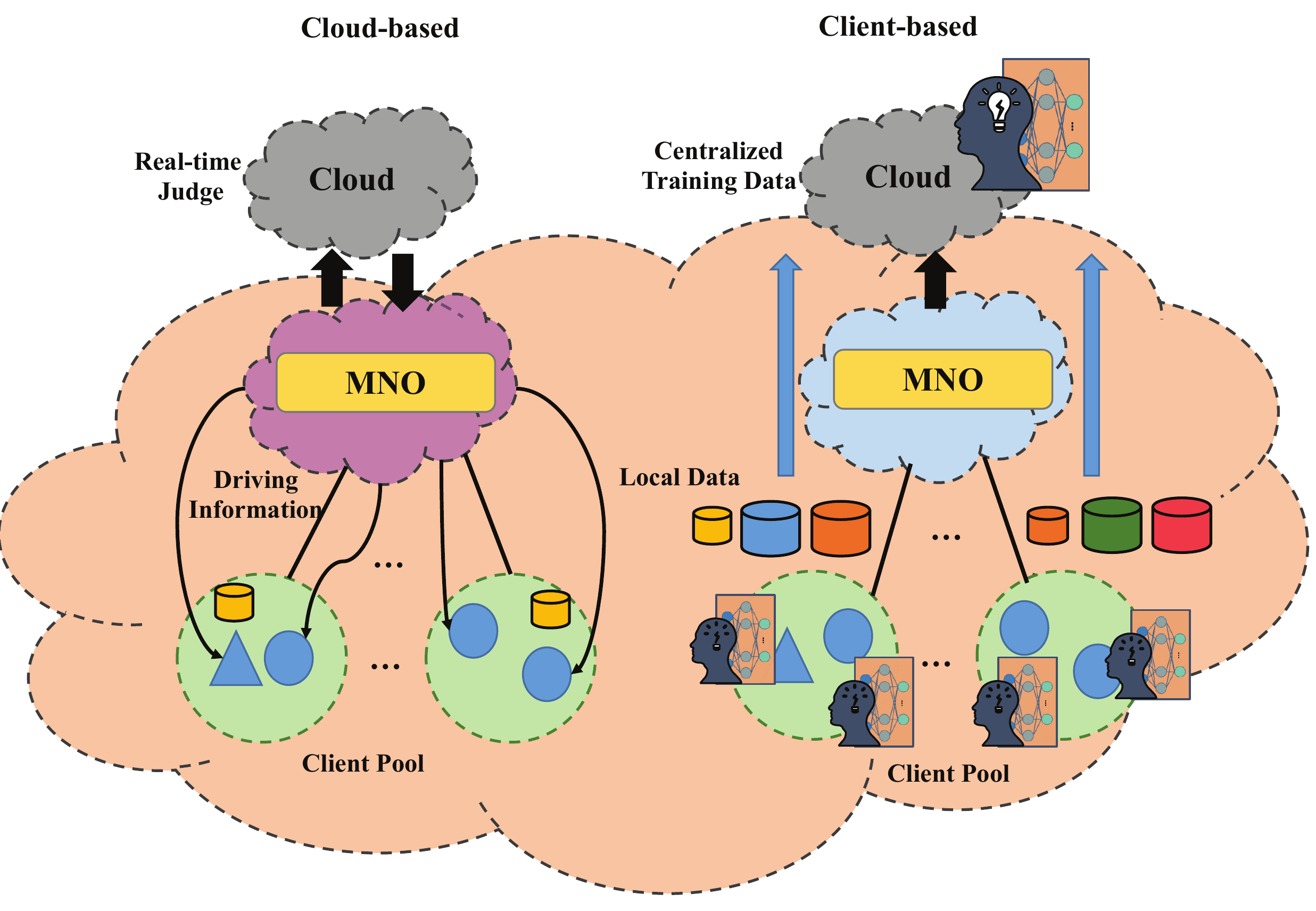}}
  \caption{Two types of existing approaches for fatigue detection. Vehicles transmit data through mobile network operators (MNOs). Traditional detection methods need massive communication costs and very sensitive to network delay.}
\end{figure} 

Many fatigue supervision methods have emerged with the popularization of intelligent equipment and the IoV. The methods judge whether the driver is fatigued by analyzing changes of the collected drivers’ facial expressions such as blinking, eye closure duration, etc. The existing work can be divided into two categories as shown in Fig. 1: i) \textit{the cloud-based method}: the server located in the cloud makes judgments by simply analyzing the uploaded driving information and informs results to clients \cite{refer2}; ii) \textit{the client-based method}: the client directly judges fatigue actions based on the implanted supervision model, which is trained in the server using the uploaded driving data \cite{refer5}. Nevertheless, these methods may suffer from the following issues: i) the cloud-based method lacks intelligence and is vulnerable to network delays owing to the transmission of driving information and judgment results; ii) the client-based method cannot ensure the accuracy of judgments because of lacking dynamic optimization of the trained supervision model; iii) the two methods pose serious security risks to clients’ privacy since data collected by cloud for centralized training.

The Federated Learning (FL) proposed by Google can solve the above-mentioned problems by training models on clients, optimizing the central model through constantly aggregating the updated local training models, at last sending the updated model to clients \cite{refer10}. Its effectiveness has been verified in some related studies such as hierarchical FL \cite{refer8,refer9}. However, the proposed FL is based on the two-level framework composed of cloud and client levels, considering vehicles as training nodes that lack data labeling capabilities and require low latency, which is not feasible for the practical IoV scenario where computing and communication resources are limited. To solve the resource limitation problem, \cite{zhongxing} proposed migrating machine learning to the edge computing scenario, which places additional computing devices at the edge to address latency and privacy problem. \cite{xiao1, xiao2} proposed a dynamic allocation algorithm of edge resources (DAERs), which improve the prediction accuracy of the staged destination of the vehicle to help allocate resources reasonably. However, none of these work considered the following problems: i) The client may not have sufficient computing resources; ii) The transmission of training data may produce large amounts of communication overhead.

\begin{figure}[ht]
  \centering
  \scalebox{0.4}{\includegraphics{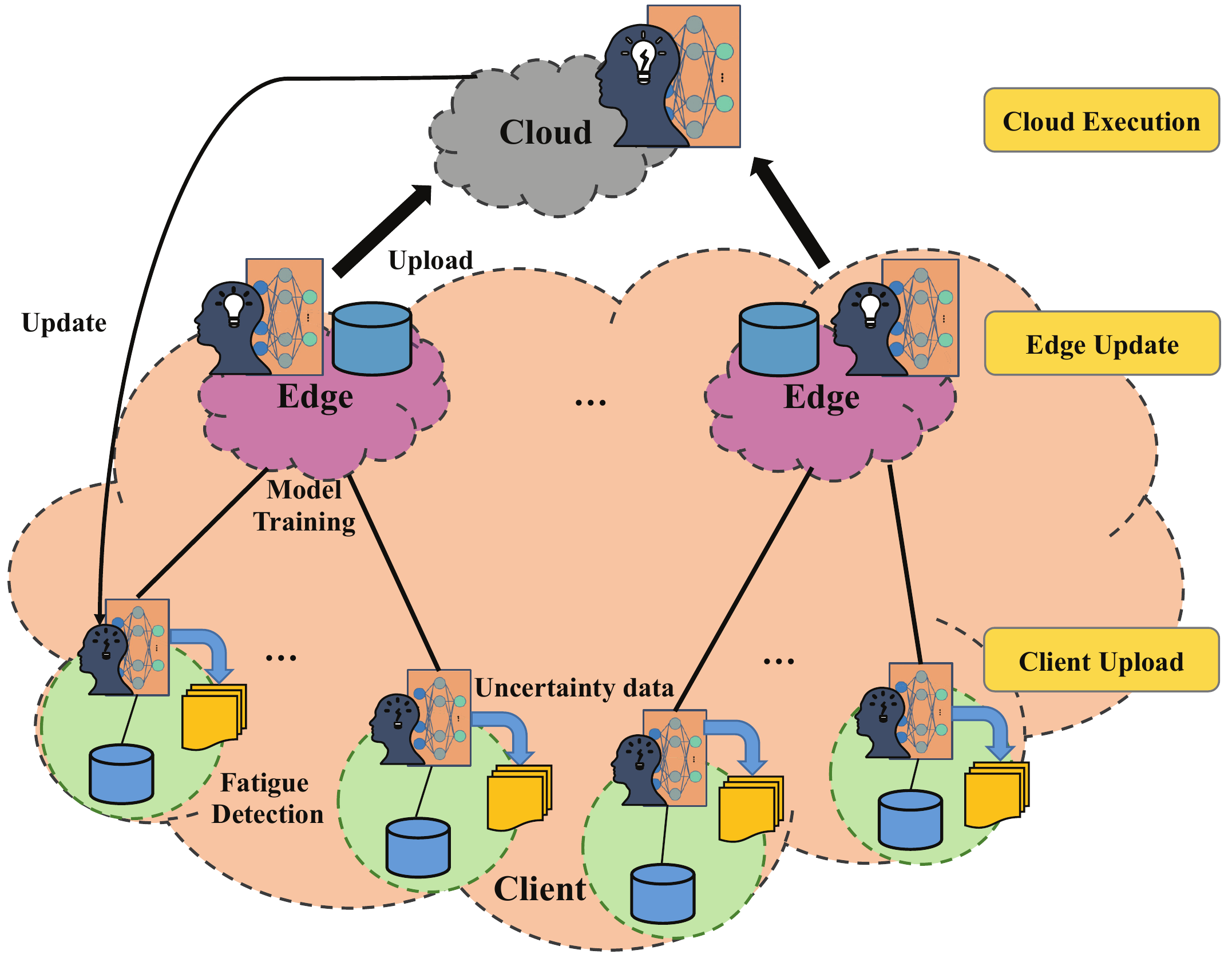}}
  \caption{The proposed FedSup framework.}
\end{figure} 

In this paper, we propose a Federated learning-based fatigue driving behaviors Supervision framework (FedSup), in which the edge node trains the local model using uncertainty data of images uploaded by clients and only uploads model parameters to the cloud server for aggregation, and the server sends the aggregated training model to clients to judge fatigue actions. On the one hand, it reduces the computation pressures of clients and communication overhead. On the other hand, edge nodes upload parameters instead of raw data to the cloud server for model optimization ensuring clients’ data privacy, and the server distributes parameters to all edge nodes and clients realizing a dynamic update of the model.

Our main contributions are illustrated as follows:
\begin{itemize}
\item We design cloud-edge-client fatigue driving supervision framework FedSup, which is communication-efficient, accurate, and privacy-preserving.

\item In FedSup, we propose an efficient method to quantify the uncertainty of the image, which minimizes the amount of uploaded data and requirements of clients’ computing resources. 

\item In FedSup, we propose a parameter aggregation algorithm based on uncertainty weight to more effectively integrate local models trained in edge nodes, which improves the accuracy of the central model and reduces communication rounds in the training process.
\end{itemize}

The remaining of this article is organized as follows. In Section II, we introduce the related work. The FedSup framework design details the proposed algorithm, especially the Bayesian approximate in convolution neural network (CNN), and uncertainty weighted aggregation are described in Section III. In Section IV, we present the experimental results of the FedSup framework on two real-world data sets. Finally, conclusions are drawn in Section V.

\section{Related Work}
This section reviews research related to our method, mainly contains two aspects, fatigue detection, and federated Learning.

\subsection{Fatigue detection}
Recently research on fatigue can be divided into three primary types: one type is based on driver’s physiological response \cite{refer11,refer12}, a detection system that uses wearable sensors to detect the driver’s alcohol content and driving posture but arouses disgust of the driver and even affects driving behavior; the second type is based on the driving data \cite{refer14}, an analysis and detection system that uses vehicle-mounted sensor equipment to detect vehicle motion characteristics; the third type is based on the driver’s facial features \cite{refer15}, the driver’s driving behavior is captured and analyzed by the in-car monitoring equipment. Among them, facial features are considered to be the most promising technology for detecting abnormal driving behavior.

After that, Zhang et al. \cite{refer17} proposed an eye status recognition algorithm to calculating the percentage of eyelid closure over the pupil over time (PERCLOS), and based on blink frequency to judge fatigue. The algorithm can effectively detect fatigue, and the PERCLOS criterion has become one of the standards for fatigue detection.

Based on the PERCLOS criterion. Xiao et al. \cite{refer18} treat surveillance videos as a set of continuous image input, detect driving status via driver eyes spatial-temporal feature, and a CNN network with long and short time memory (LSTM) units is designed.

Each of these methods has its own merit. However, these approaches lack a data-sharing mechanism to protect privacy, and it is still improvable in communication cost.

\subsection{Federated Learning}

As elaborated above, up to now, model training in IoV is privacy-exposed and non-dynamically. With the emergence of federated learning and edge computing, such a problem is expected to be solved. The detail about representative exiting FL works as follows.

Brendan et al. \cite{refer10} propose a practical technique called federated learning of deep networks based on iterative model averaging. The approach is robust to the unbalanced and non-IID data distributions, and federated averaging (FedAVG) has also become the benchmark aggregation algorithm in FL.

\begin{figure*}[htbp]
  \centering
  \scalebox{0.5}{\includegraphics{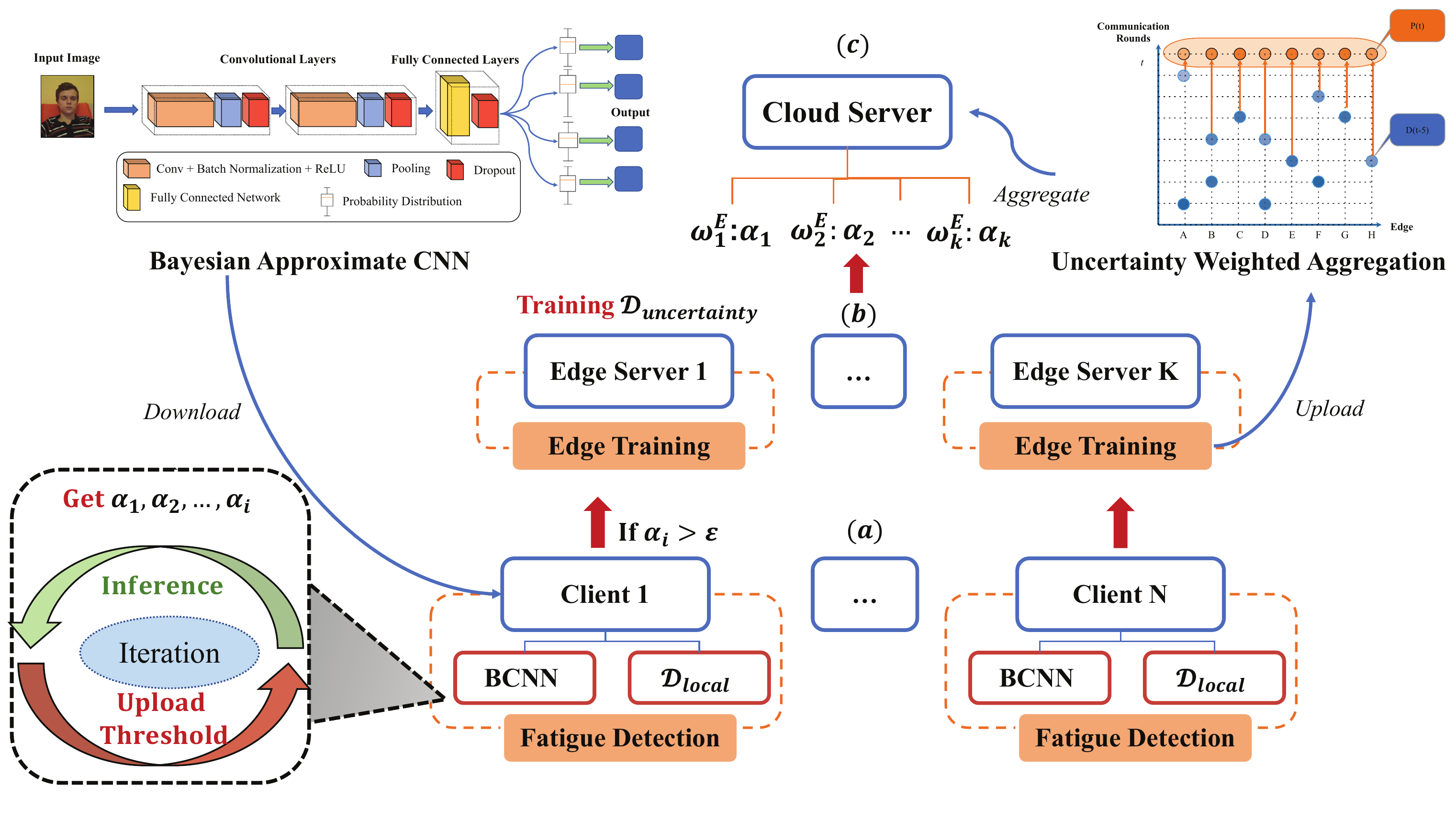}}
  \caption{Overview of the FedSup framework, which consists of three parts, Client Upload: uncertainty data upload, Edge Update: model training, and Cloud Execution: weighted aggregation.}
\end{figure*} 

Inspired by the FL technique, Andrew et al. \cite{refer19} utilizes a recurrent neural network (RNN) for language model training using federated learning for next-word prediction. This work proves the feasibility and advantages of applying federated learning to smart mobile devices for training. Such practice is only suitable for the scenario in which the client-side can annotate data during use.

Aiming at reducing communication overhead of parameter passing. Chen et al. \cite{refer20} presents an asynchronous learning strategy on the clients and a temporally weighted aggregation of the server’s local models. The parameters of the deep layers are updated less frequently than those of the shallow layers. However, this method is not suitable for neural networks with a few layers.

Wang et al. \cite{refer21} propose a control algorithm that determines the best tradeoff between local update and global parameter aggregation to minimize the loss function under a given resource budget. However, this algorithm is only suitable for scenarios that computing and communication resources are determined.

In this paper, we first apply the combination of model uncertainty and FL to fatigue driving detection. Dynamically improve the accuracy of the central model under the premise of maximizing user privacy.

\begin{table}[]
  \caption{List of Meta-parameters}
  \setlength{\tabcolsep}{5mm}
  \begin{tabular}{ccc}
      \toprule  
      \textbf{Name}& \textbf{Description} \\
      \midrule 
      $\omega^C$& Cloud model parameters  \\
      $\omega^E$& Edge model parameters  \\
      $\varepsilon$& Upload threshold of image uncertainty value for client  \\
      $\alpha$& uncertainty value   \\
      \bottomrule
      \end{tabular}
\end{table}

\section{FedSup Design}
As shown in Fig. 3, the FedSup framework consists of three parts denoted as Client Upload, Edge Update, and Cloud Execution. Initially, in the Client Upload part, clients use the local model for fatigue detection while \textbf{quantifying the uncertainty of image data}, then uploading the data to a nearby edge server. In the Edge Update part, the edge server utilizes uncertainty data for model training and update parameters to the cloud for aggregation. Finally, in the Cloud Execution part, aiming at the updated model for all participants, cloud server aggregate model parameters with \textbf{uncertainty weighted aggregation algorithm}.

The details of the three parts will be introduced in this section. Table I lists the meta-parameters of the FedSup framework.

\subsection{Client Upload}
Deep learning tools have gained tremendous attention in edge computing. However, such tools do not capture model uncertainty. Bayesian probability theory offers us mathematically grounded methods to quantify model uncertainty while usually come with a prohibitive computational cost. Thus, we perform dropout training in the CNN as an approximate Bayesian inference.

Our idea was mainly inspired by the following observations on CNNs. 1) Training data sets with different uncertainty values have different contributions to the model. 2) Training data sets with higher uncertainty usually more helpful for model optimization. Our uncertainty data upload method achieves uploading 1 or even 2 orders of magnitude fewer data in learning iteration.

\begin{figure*}[ht]
  \centering
  \scalebox{0.5}{\includegraphics{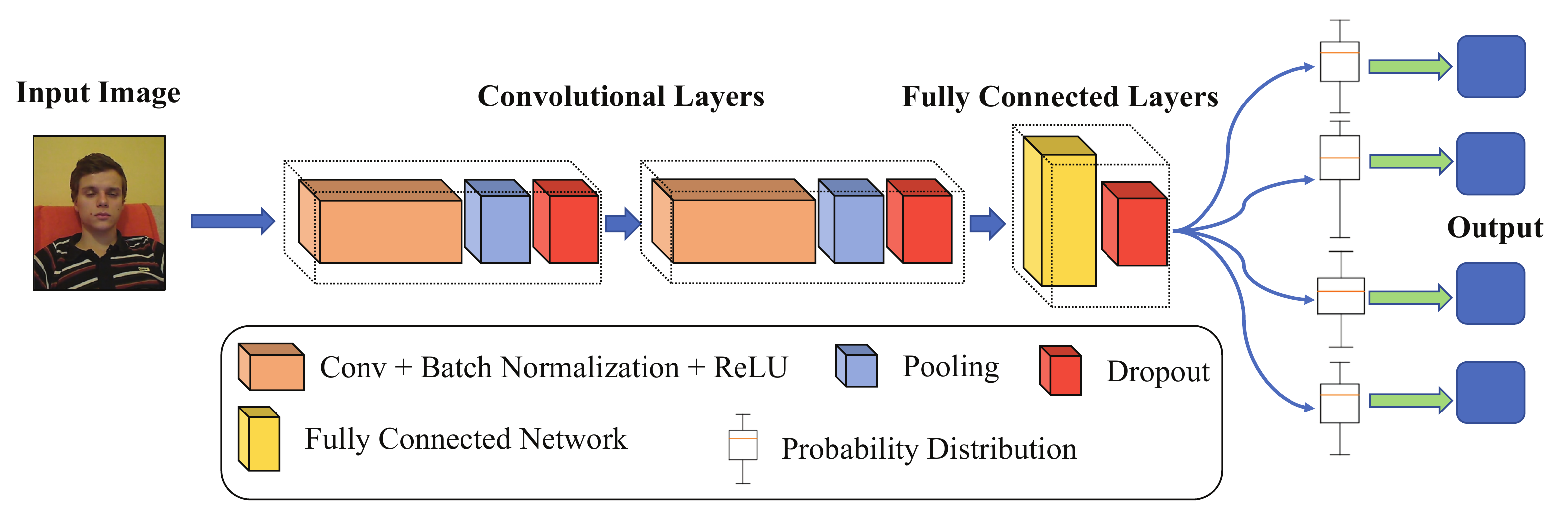}}
  \caption{Schematic illustrations of the proposed fatigue detection method.}
\end{figure*} 

We now discuss the methods of fatigue detection and uncertainty analysis. During driving, the continuous videos are collected by the vehicle’s camera as input. For each input frame, we extract the eyes feature, and the driving status is determined based on the number of consecutive frames of eyes blink.

1) Eye Landmarks: Eye landmarks detection is a fundamental component in FedSup. It aims to localize eye corners. The algorithm theory that we used is based on \cite{refer18}, which uses CNN in facial landmarks recognition. 

2) Eye Feature Extraction: According to the algorithm theory of \cite{refer22}, we combine the Gabor and Local binary patterns (LBP) features to judge the eye blink state. The Gabor and LBP feature maps are used as input data to pretrain the CNN network. When pretraining is completed, the original image is used as input to train the model, the driver’s driving state is judged by the PERCLOSE criterion \cite{refer23}.

The uncertainty analysis of approximate posterior marginalization through Monte Carlo integration, such as
\begin{equation}
p(y=c|D_{train})= \int p(y=c|x,\omega)p(\omega|D_{train})d\omega, \label{} 
\end{equation}
where $x$ is input image in this round, $c$ is category, and $\omega$ is model parameters.

In FedSup, as shown in Fig. 4, the Bayesian convolutional neural network framework is implemented by utilizes dropout layers at both training and inferencing phases \cite{refer24} to get the posterior distribution. The result’s mean and variance are used as confidence and uncertainty, respectively \cite{refer25}. Here, we set
\begin{equation}
  \begin{aligned}
r = \frac{1}{M} \sum^M_{m=1}p(y=c|x,\widehat{\omega}_t),\\ 
\alpha = \frac{1}{M} \sum^M_{m=1}[r-p(y=c|x,\widehat{\omega}_t)]^2 \label{}
  \end{aligned}
\end{equation}
where $\widehat{\omega}_t$ is multiple detection model parameters with dropout, $M$ indicates how many BCNN the data will pass through to get the result, $r$ means the classification result of the network output, $\alpha$ means the uncertainty of the image data. If $\alpha$ is higher than $\varepsilon$, the image will upload to the edge server for training.

\begin{algorithm}
  \renewcommand{\thealgorithm}{1}
  \caption{Client Component of FedSup. The $M$ BCNN are indexed by $m$.}
  \begin{algorithmic}[1]
      \State{//Run on Client $n$}
      \Function{ClientUpload}{$n,\omega$}
      \State{Initialization local $\omega\leftarrow \omega^C$}
      \For{each image data $i$=1,2,...}
          \For{each neural network $m$ from 1 to $M$}
              \State{$r_m,\alpha_m\leftarrow$ BCNN($d_i$)}
          \EndFor
          \State{$r \leftarrow $ Average($r_1,...r_m$)}
          \State{$\alpha_i \leftarrow $ Variance($\alpha_1,...\alpha_m$)}
          \If{$\alpha_i \geq \varepsilon$}
              \State{$dict(d_i)\leftarrow \alpha_i $}
          \EndIf
      \EndFor
      \\ \Return{$dict$}
      \EndFunction
  
\end{algorithmic}
\end{algorithm}

The implementation of the Client Upload (Algorithm 1) is controlled by $n$, $M$ and $\varepsilon$, where $n$ is the selected clients’ index, $\varepsilon$ indicates the upload threshold of image uncertainty value for the client. In Algorithm 1, lines 4-9 perform real-time inference through the BCNN and quantify the data’s uncertainty value, lines 10-12 judge whether the data is uploaded to the edge server, line 14 return uncertainty data dictionary. 

\begin{algorithm}[htbp]
  \renewcommand{\thealgorithm}{2}
  \caption{Edge Server Component of FedSup. The $N$ client are indexed by $n$, $E$ is local training numbers.}
  \begin{algorithmic}[1]
      \State{//Run on Edge $k$}
      \Function{EdgeUpdate}{$k,\omega$}
          \State{Download $\omega^C$ from Cloud Server and replace the corresponding local $\omega$}
          \State{Input$\leftarrow$ClientUpload(n)}
          \For{each local epoch $t$ from 1 to $E$}
          \State{$\omega^k_{t+1} \leftarrow \omega^k_t-\eta\bigtriangledown l(w;b)$}
          \EndFor
          \State{$\alpha^E \leftarrow $Averege$(\alpha_1,...,\alpha_n)$}
          \\ \Return{$\alpha^E, \omega^k_{t+1}$}
      \EndFunction
\end{algorithmic}
\end{algorithm}

\begin{figure}[htbp]
  \centering
  \scalebox{0.35}{\includegraphics{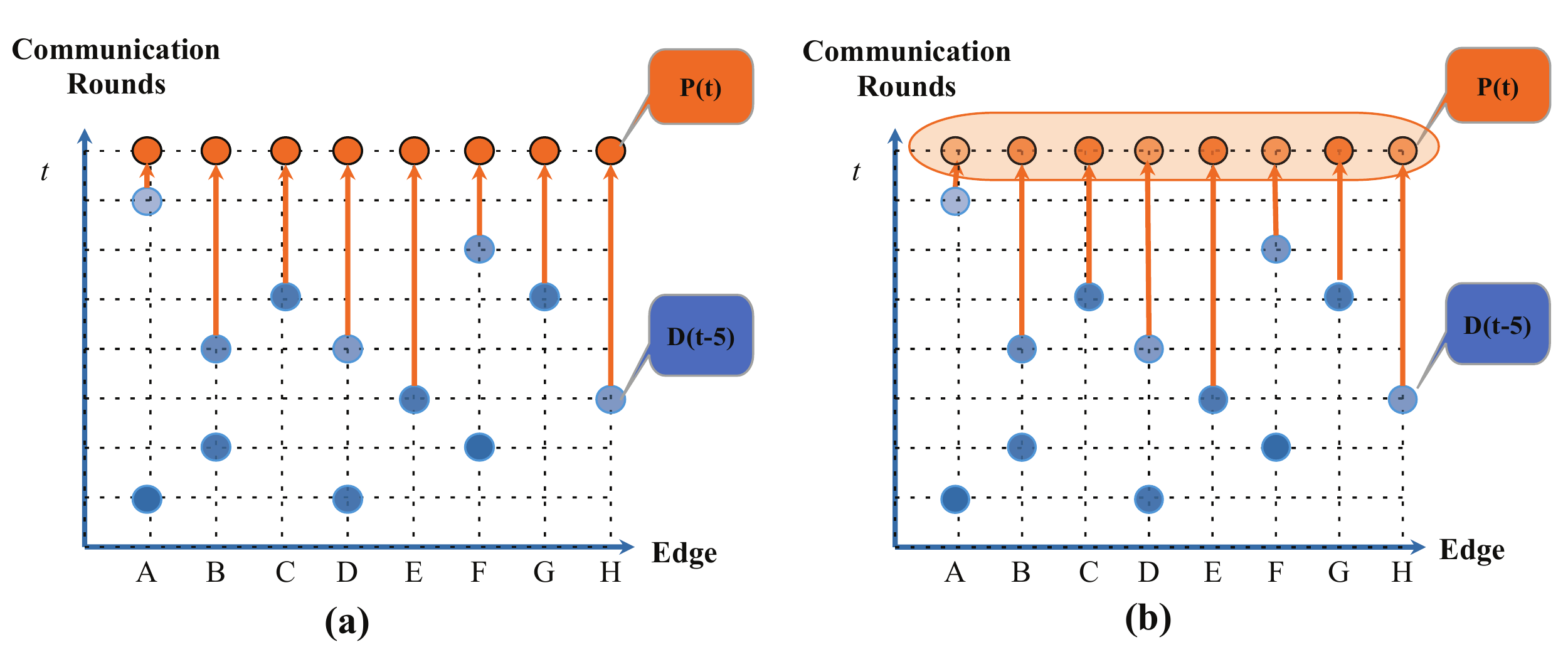}}
  \caption{(a) Conventional aggregation strategy, weighted aggregation may not be reasonably weighted only by edge data size. We evaluate the loss function from pre-trained CNN and found that with updated, the uncertainty value of edge data is constantly changing, which inspired us. (b) Illustration of the Uncertainty Weighted Aggregation. The updated model parameters are computed by Eq. (3).}
\end{figure} 

\subsection{Edge Update}
The Edge Update (Algorithm 2) is mainly composed of local training and parameters update. Line 3 edge server download newest $\omega^C$ as local model parameters, subfunction Client Upload is called to get client’s uncertainty data, $\eta$ is the learning rate. In line 6, the local training is performed to update $\omega^k_{t+1}$, $\alpha^E$ denote the average uncertainty value of all uncertainty data. Line 9 return $\omega^k_{t+1}$ and $\alpha^E$.

\subsection{Cloud Execution}
In FL, the aggregation algorithm always weights the importance of each participant by data size \cite{refer10}, that is, the larger $n^k$ is, the participant $k$ will be more valuable to the optimization of the central model. As shown in Fig. 5(a), blue dots represent the edge server’s local data set, orange dots represent the model parameters after local training. All participants’ model parameters will be weighted by local data size only, which might not be reasonable.

In FL, however, the data size and uncertainty of each edge server are changing with the central model optimization. Therefore, the edge with higher uncertainty training data set should have a larger weight in the aggregation. Accordingly, the uncertainty weighted asynchronous aggregation (UWAA\_FedAVG) is proposed, that is,
\begin{equation}
\omega_{t+1}\leftarrow \sum^K_{k=1}e^{\alpha_k} \frac{n_k}{n} \omega^k_{t+1}, \label{}
\end{equation}
where $t$ is current aggregation rounds, $\alpha_k$ means the value of data uncertainty at the edge, $\omega^k_{t+1}$ means the newest model of the $k$-th edge. The proposed UWAA\_FedAVG is shown in Fig. 5(b). Similar to the weighting method in Fig. 5(a), the orange dots represent the edge model parameters after local training, and the color shade represents the uncertainty value. The higher uncertainty of the edge data set is, the larger weight the current aggregation will have.

Cloud server (Algorithm 3) mainly implements aggregation algorithm, including initialization steps and several communication rounds. Lines 4 randomly selects subsets $C$ as part of the training from edge servers. In lines 5-7, subfunction Edge Update is called for update central model parameters $\omega^k_{t+1}$. In line 8, the UWAA\_FedAVG is performed to get $\omega_{t+1}$.

\begin{algorithm}[htbp]
  \renewcommand{\thealgorithm}{3}
  \caption{Cloud Server Component of FedSup. The $K$ Edge Servers are indexed by $k$, $C$ is the fraction of Edge Servers that perform training each round.}
  \begin{algorithmic}[1]
      \Function{CloudExecution}{}
      \State{initialize $\omega^C$}
      \For {iterations $i$=1,2,...}
          \State{$C \leftarrow$ randomly selected edge set}
          \For {each Edge $k \in$ $C$ in parallek}
              \State{$\omega^k_{t+1},\alpha_k\leftarrow$EdgeUpdate$(k,\omega^C)$}
          \EndFor
          \State{$\omega_{t+1}\leftarrow \sum^K_{k=1}e^{\alpha_k} \frac{n_k}{n} \omega^k_{t+1}$}
      \EndFor
      \EndFunction
\end{algorithmic}
\end{algorithm}

\section{Evaluation Results}
We conduct experiments on communication costs and accuracy, respectively. The experiment design, data sets, and results analysis are divided into the following three subsections.
\subsection{Experimental Design}

We evaluated the FedSup framework perform on two real-world data sets, ZJU data set \cite{refer26}, and Eyeblink8 \cite{refer27} data set, the most popular data sets for learning eye blink status. The two data sets are in unstructured videos, similar to the original fragmented data in the IoV scenario.

1) ZJU data set: This data set contains 80 videos, each of these lasting a few seconds. 20 volunteers in four state types (frontal view with and without glasses and upward view), videos refresh rate is 30fps with resolution 320x240.

2) Eyeblink8 data set: This data set contains 8 videos with four volunteers. Videos are recorded in an indoor scenario, similar to the Talking face data set. There is 408 eye blinks on 70 992 annotated frames with resolution 640x480.

FedAVG \cite{refer10} has chosen as the benchmark algorithm since it is the SOAT FL approach. The UWAA\_FedAVG is also compared with two traditional methods, that is, Centralized\_SGD that collect all data for centralized training, and Standalone\_SGD, which trains local data sets on each edge side without sharing, because they are the mainstream training methods for fatigue driving in the existing IoV.

\begin{table}[htbp]
  \caption{NN architecture for the eye landmarks CNN}
  \setlength{\tabcolsep}{12mm}
  \begin{tabular}{ccc}
    \toprule  
    \textbf{Layer}& \textbf{Shapes} \\
    \midrule  
    Conv2d\_1& 3 $\times$ 3 $\times$ 32 \\
    Maxpool\_1+Dropout& 2 \\
    Conv2d\_2& 3 $\times$ 3 $\times$ 64 \\
    Maxpool\_2& 2 \\
    Conv2d\_3& 2 $\times$ 2 $\times$ 128 \\
    Dense\_1+Dropout& 1024 \\
    Dense\_2+Dropout& 1024 \\
    Dense\_3& 10 \\
    \bottomrule 
  \end{tabular}
\end{table}

\begin{table}[htbp]
  \caption{NN architecture for the eye feature extraction CNN}
  \setlength{\tabcolsep}{12mm}
  \begin{tabular}{ccc}
    \toprule  
    \textbf{Layer}& \textbf{Shapes} \\
    \midrule  
    Conv2d\_1& 3 $\times$ 3 $\times$ 32 \\
    Maxpool\_1+Dropout& 2 \\
    Conv2d\_2& 3 $\times$ 3 $\times$ 64 \\
    Maxpool\_2& 2 \\
    Dense\_1+Dropout& 128 \\
    Dense\_2& 2 \\
    \bottomrule 
  \end{tabular}
\end{table}

\subsection{Data Sets}
As mentioned in Section I, IoV scenarios has its own special challenges, such as data distributed and unbalanced. To simulate the distributed applications and the IoV scenarios with multiple participants, we divided each data set into unbalanced parts based on the number of edges and clients. Thus, a client may hold several data with different people and scenes, which increases the training difficulty.

Each data in ZJU is a 320$\times$240-pixel RGB image. We divide the data into 50 unequal parts. The size of client data obeys a normal distribution with an expectation of 400 and a variance of 10, 20, and 30. Namely, 10@ZJU, 20@ZJU, and 30@ZJU are predefined.

The eye landmarks CNN architecture with two 3$\times$3 convolution layers (the first with 32 channels and the second with 64), each following with 2$\times$2 max-pooling layers, one 2$\times$2 convolution layers (with 128 channels). Two fully connected layers (with 512 and 1024 units) and ReLu activation, and a softmax output layer. Table II shows the neural network (NN) architecture.

Each data in Eyeblink8 is a 640$\times$480-pixel RGB image. We divide the data into 50 unequal parts. Here, The size of client data obeys a normal distribution with an expectation of 1400 and a variance of 50, 100, and 150. namely, 50@ Eyeblink8, 100@ Eyeblink8, 150@Eyeblink8 are predefined.

The eye feature extraction CNN architecture with two 3$\times$3 convolution layers (the first with 32 channels and the second with 64), each following with 2$\times$2 max-pooling layers, a fully connected layers with 512 units and ReLu activation, and a softmax output layer. Table III shows the neural network architecture.

Since the FedSup framework utilizes the dropout mechanism to construct the Bayesian approximate CNN, the first convolutional layer and the fully connected layer are followed by a dropout layer \cite{refer25} to quantify the model uncertainty.

\subsection{Results and Analysis}
In this subsection, we will carry out two sets of experiments around the FedSup framework to verify its effectiveness. The first part evaluates the impact of key parameters $\varepsilon$, $M$, $K$, and $N$ in the FedSup framework on performance, and the second part compares the framework with four algorithms in terms of communication overhead and accuracy.

\subsubsection{Effect of Key Parameters}
Experiments with key parameters may helpful for us to understand the performance of FedSup framework in IoV environment. When evaluating one key parameter, other parameters set default values. Experiment on $\varepsilon$ = \{0.02, 0.025, 0.03\}, given $C=0.3$, $K=10$, $N=50$, and $M=3$.

\begin{table}[htbp]
  \caption{Experimental Results on $\varepsilon$}
  \setlength{\tabcolsep}{7.5mm}
  \begin{threeparttable}
  \centering 
  \begin{tabular}{ccc}
  \toprule 
  \textbf{$\varepsilon$}& \textbf{Accuracy\tnote{*}}& \textbf{Rounds\tnote{**}} \\
  \midrule
  0.020& 91.23\%(0.0042) & 173.2(21.66)\\
  0.025& 90.74\%(0.0068) & 155.3(33.12)\\
  0.030& 90.66\%(0.0195) & 151.7(34.79)\\
  
  \bottomrule 
  \end{tabular}
  
  \begin{tablenotes}
      \footnotesize
      \item[*] Average(Standard Deviation) of best accuracy within 200 training epoch.
      \item[**] Communication rounds are required for UWAA\_FedAVG to achieve 90.00\% accuracy on 50@Eyeblink8.   
  \end{tablenotes}
  \end{threeparttable}
  \end{table}

\begin{table}[htbp]
  \caption{Experimental Results on $M$}
  \setlength{\tabcolsep}{8mm}
  \begin{threeparttable}
  \centering 
  \begin{tabular}{ccc}
  \toprule  
  \textbf{$M$}& \textbf{Accuracy\tnote{*}}& \textbf{Rounds\tnote{**}} \\
  \midrule  
  3& 90.41\%(0.0082) & 187.2(22.34)\\
  5& 90.74\%(0.0094) & 161.8(31.37)\\
  7& 91.17\%(0.0067) & 152.4(25.82)\\
  
  \bottomrule 
  \end{tabular}
  
  \begin{tablenotes}
      \footnotesize
      \item[*] Average(Standard Deviation) of best accuracy within 200 training epoch.
      \item[**] Communication rounds are required for UWAA\_FedAVG to achieve 90.00\% accuracy on 50@Eyeblink8.  
  \end{tablenotes}
  \end{threeparttable}
\end{table}

\begin{table}[htbp]
  \caption{Experimental Results on $N$ and $K$ of FedSup framework}
  \setlength{\tabcolsep}{5.3mm}
  \begin{threeparttable}
  \centering 
      \begin{tabular}{cccc}
      \toprule  
      \multicolumn{2}{l}{\textbf{Scalability}} & \multirow{2}{*}{\textbf{FedAVG}} & \multicolumn{1}{r}{\multirow{2}{*}{\textbf{UWAA\_FedAVG}}} \\ \cmidrule(r){1-2}
      \textbf{$K$}& \textbf{$N$}& & \multicolumn{1}{r}{} \\ \midrule 
      10&20&89.91\%(0.0063)&\textbf{90.32\%(0.0064)}  \\
      10&40&90.16\%(0.0091)&\textbf{90.46\%(0.0033)}  \\
      10&60&90.25\%(0.0058)&\textbf{90.40\%(0.0047)} \\
      15&30&\textbf{90.31\%(0.0032)}&90.23\%(0.0062)  \\
      15&45&\textbf{90.47\%(0.0055)}&90.46\%(0.0045) \\
      15&60&90.45\%(0.0113)&\textbf{91.16\%(0.0037)}  \\
      20&40&90.75\%(0.0064)&\textbf{90.91\%(0.0049)}  \\
      20&60&90.73\%(0.0042)&\textbf{91.04\%(0.0102)} \\
      20&80&91.26\%(0.0056)&\textbf{92.11\%(0.0093)}  \\
      \bottomrule 
      \end{tabular}
  \end{threeparttable}
\end{table}
We adopted two metrics to measure UWAA\_FedAVG performance, the best test accuracy of the model within 200 training epoch; the other is communication rounds required before model accuracy reaches 90.00\%. Each experiment running ten times independently. Table IV-VI shows the average and standard deviation of the model accuracy and communication rounds for training.

\begin{figure}[htbp]
  \centering
  \scalebox{0.3}{\includegraphics{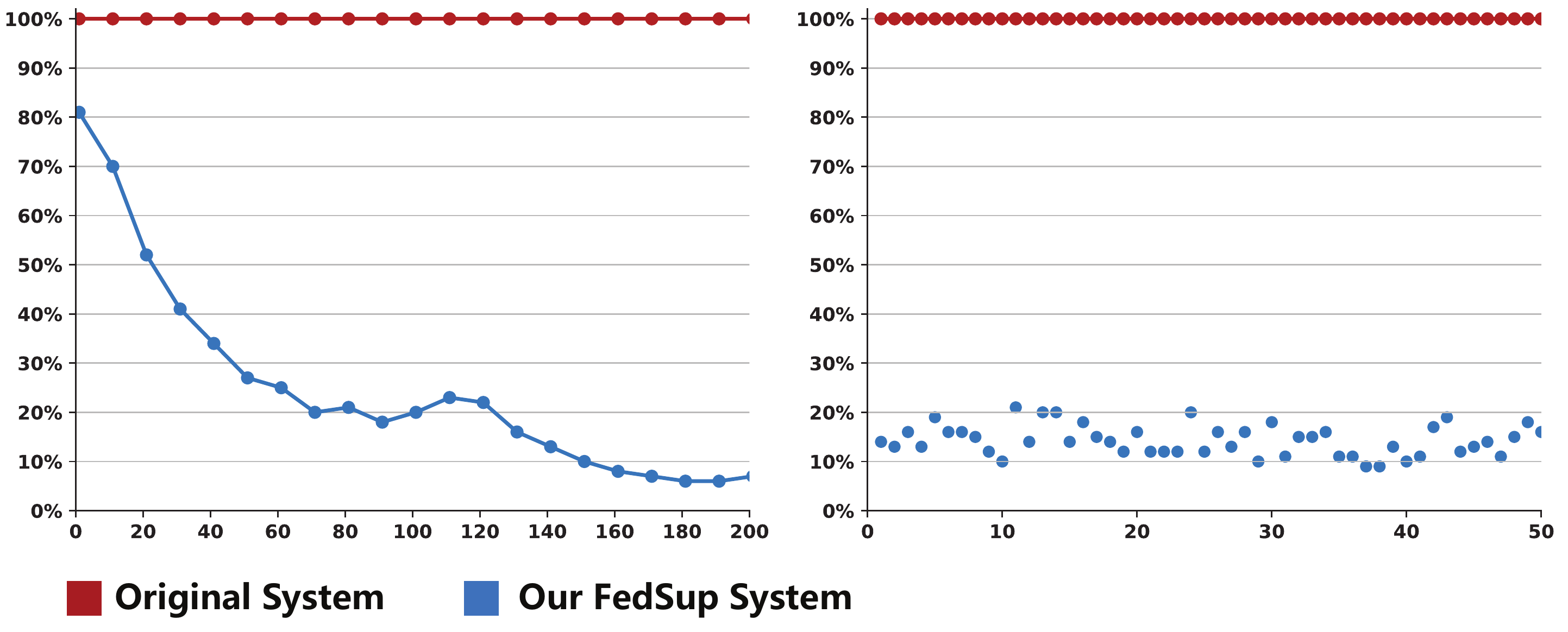}}
  \caption{Left: The Client Upload Data Amount Line Chart displays the average uploaded data amount (ratio of uploaded to total data) by the client in 200 iterations during the training process. Right: The Client Upload Data Amount Scatter Chart displays the average uploaded data amount by 50 clients during the optimize process.}
\end{figure} 

\begin{figure}[htbp]
  \centering
  \scalebox{0.4}{\includegraphics{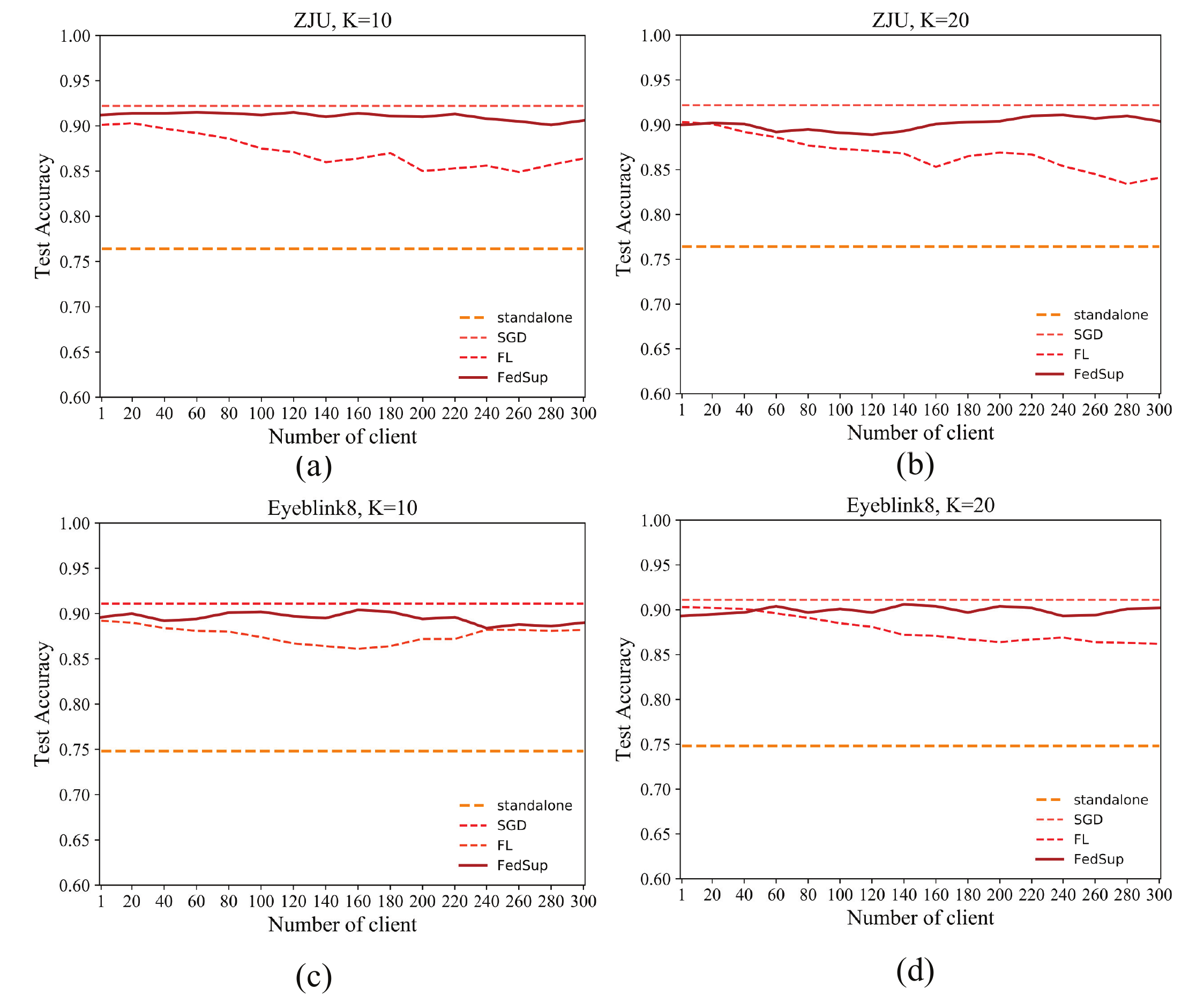}}
  \caption{Optimal model accuracy under different edge and client numbers. The x-axis is the client number, (a)-(d) are the impact of different client numbers corresponding to the edge server.}
\end{figure} 

Based on the results in Table IV-VI, we can draw the following conclusions.
\begin{itemize}
  \item Experiment on $\varepsilon$: $\varepsilon$ is upload threshold of uncertainty value for the client, based on the results show in Table IV, we can conclude that the higher upload threshold is, the higher accuracy will get, and fewer communication rounds will be required, that is expected. A higher threshold can help the model filter more valuable data for training. However, a too higher threshold will reduce the training data size and make the model underfit.

  \item Experiment on $M$: $M$ is the number of CNN with dropout which client inference. The result in Table V indicates that the larger $M$ is, the lower communication rounds will be required, and the higher model accuracy will get, but this will increase the computational burden during client detection when $M$ =1 and $\varepsilon=0$, the algorithm is reduced to FedAVG, meaning that client upload all data to edge server.
\end{itemize}

\begin{figure}[htbp]
  \centering
  \scalebox{0.7}{\includegraphics{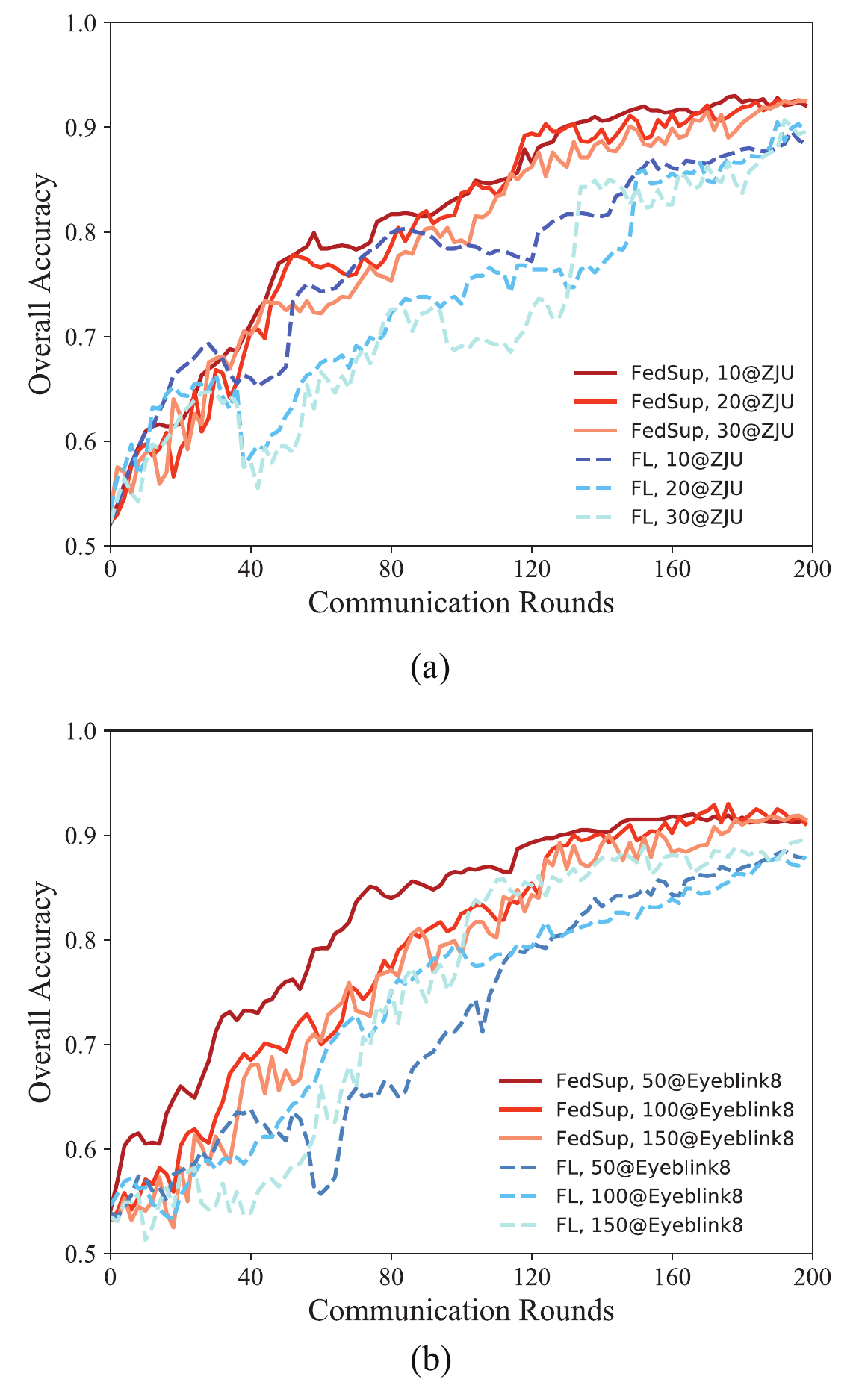}}
  \caption{Comparison of model accuracy between UWAA\_FedAVG and FedAVG algorithm on ZJU (a) and Eyeblink8 (b) data sets.}
\end{figure} 

Considering the instability of the connection state in the IoV scenario, we demonstrated the FedSup framework’s scalability by adjusting the number of edge servers $N$ and the number of clients $K$. Based on the results show in Table VI, we can draw the following conclusions.
\begin{itemize}
  \item The more participants in the FedSup framework, the better for central model training, and the FedSup framework performance better than the baseline algorithm in most cases.
  \item The larger the client number corresponding to the edge server, the accuracy of the FedSup framework will be higher, which aligns with the actual situation that each edge in IoV may link multiple vehicles.
\end{itemize}

\subsubsection{Comparison with baseline algorithm}
The following experiments compare the FedSup framework with other SOAT methods in terms of communication overhead and model accuracy performance. The baseline algorithm includes FedAVG, Centralized\_SGD, and Standalone\_SGD. Figs. 6-8 shows the comparison results on ZJU and Eyeblink8 data sets.

To evaluate the FedSup framework‘s advantages in terms of communication overhead, we compared the amount of client data upload from two aspects: i) For 200 rounds training process; ii) For 50 clients. The experimental results, as shown in Fig. 6, the average uploaded amount of our systems has reduced by about 85\% compared to the baseline system in both two situations.

\begin{table}[ht]
  \caption{Experimental on Performance}
  \setlength{\tabcolsep}{3.1mm}
  \begin{threeparttable}
  \centering 
      \begin{tabular}{ccccc}
      \toprule 
      \multirow{2}{*}{\textbf{ID@data set}} & \multicolumn{2}{c}{\textbf{FedAVG}} & \multicolumn{2}{c}{\textbf{UWAA\_FedAVG}} \\ 
      \cmidrule(r){2-3} \cmidrule(r){4-5} 
       & \textbf{Accuracy\tnote{$\star$}}& \textbf{Rounds}& \textbf{Accuracy\tnote{$\star$}}& \textbf{Rounds}\\ \hline
      10@ZJU\tnote{*}& \textbf{90.35}\%& 174& \multicolumn{1}{c}{90.24\%}&  \textbf{157}\\
      20@ZJU\tnote{*}& 90.64\%& 197& \multicolumn{1}{c}{\textbf{92.96\%}}& \textbf{146}\\ 
      30@ZJU\tnote{*}& 90.61\%& 226& \multicolumn{1}{c}{\textbf{91.44\%}}& \textbf{172}\\ 
      50@Eyeblink8\tnote{*}& 88.24\%& 256& \multicolumn{1}{c}{\textbf{90.82\%}}& \textbf{174}\\ 
      100@Eyeblink8\tnote{*}& 88.47\%& 235& \multicolumn{1}{c}{\textbf{90.27\%}}& \textbf{186}\\ 
      150@Eyeblink8\tnote{*}& 87.25\%& 278& \multicolumn{1}{c}{\textbf{90.13\%}}& \textbf{206}\\ 
      \bottomrule 
      \end{tabular}

  \begin{tablenotes}
      \footnotesize
      \item[*] Perdefined three local ZJU/Eyeblink8 data sets.
      \item[$\star$] Accuracy reaches the best 93/91\% within 200/300 rounds.
  \end{tablenotes}
  \end{threeparttable}
  \end{table}

Fig. 7 presents the best accuracy we obtain in 300 rounds when running FedSup on ZJU and Eyeblink8 for different edge and client numbers. In each plot, we show the best accuracy for centralized and standalone SGD. Both of them are independent of the a-axis since there is no parameter sharing. Comparing the accuracy of FedSup with the baselines reflects the tradeoff between utility and privacy. Our results show that FedSup using UWAA\_FedAVG with higher accuracy than FL and standalone learning, almost equal to centralized SGD.

We also observe that the client number has a lower impact on accuracy than the edge number. This indicates that FedSup does not require so much client to boost accuracy.

The following conclusion can be made based on the results show in Fig. 8. FedSup outperforms FedAVG in both data sets. The more unbalanced distribution of each client, the higher the accuracy of FedSup will reach, which demonstrates that FedSup is more suitable for IoV scenarios with unbalanced data distribution. In the ZJU data set, FedAVG has higher accuracy in the initial training period, which indicates that FedAVG performs better when the data is evenly distributed. Compared with the FedAVG model, FedSup has less fluctuation and achieves the optimal accuracy of nearly 160 rounds, while FedAVG requires more than 200 rounds, compared with a 30\% reduction in communication cost. In the Eyeblink8 data set, FedSup achieves the optimal accuracy at nearly 140 rounds, which is better than FedAVG with 190 rounds and reduces the communication cost by about 35\%.

Finally, Table VII shows the model accuracy and communication cost of the UAWW\_FedAVG and FedAVG algorithms under six data distribution cases. The following conclusions can be made. 

\begin{itemize}
\item UAWW\_FedAVG outperforms FedAVG in most cases in terms of communication rounds and best accuracy.
\item UAWW\_FedAVG utilizes uncertainty data to accelerate the central model’s convergence speed and improve the accuracy, which is more suitable for the IoV scenario with limited communication resources.
\end{itemize}

\section{Conclusion}
In this paper, we proposed FedSup, an asynchronous federated learning-based framework for fatigue detection in IoV scenarios. Experimental studies on the ZJU and Eyeblink8 data set demonstrate that our FedSup outperforms other SOAT methods in terms of communication cost and accuracy.

This paper utilizes epistemic uncertainty to quantify data uncertainty and design the UWAA\_FedAVG. In future research, we will develop a new uncertainty analysis algorithm to improve the efficiency of model aggregation further.


\footnotesize 
\bibliographystyle{IEEEtran}
\bibliography{paper}

\end{document}